\def\year{2021}\relax
\documentclass[letterpaper]{article} 
\usepackage{aaai21}  
\usepackage{times}  
\usepackage{helvet} 
\usepackage{courier}  
\usepackage[hyphens]{url}  
\usepackage{graphicx} 
\usepackage{natbib}  
\usepackage{caption} 
\usepackage{comment}
\usepackage{subcaption}
\usepackage{amsmath}
\usepackage{booktabs}
\usepackage{multirow}
\newcommand{\sign}{\mbox{sign}}

\usepackage{color}
\usepackage{xcolor} 
\newcommand{\attilio}[1]{}
\newcommand{\enzo}[1]{\textcolor{blue}{Enzo: #1}}

\usepackage{etoolbox}
\newtoggle{showrevisions}
\togglefalse{showrevisions}
\iftoggle{showrevisions}{%
\usepackage{xcolor}
\usepackage[normalem]{ulem}
\newcommand{\removed}[1]{\removedfragile{#1}}
\newcommand{\removedfragile}[1]{{\color{red}{\sout{#1}}}}

}{%
  \newcommand{\removed}[1]{} 
  \newcommand{\removedfragile}[1]{}

}

\title{LOss-Based SensiTivity rEgulaRization:\\ Towards Deep Sparse Neural Networks}

\author{
    Enzo Tartaglione, Andrea Bragagnolo, Attilio Fiandrotti and Marco Grangetto\\
}
\affiliations{
    University of Torino, Italy
}

\begin{document}

\maketitle

\begin{abstract}

LOBSTER (LOss-Based SensiTivity rEgulaRization) is a method for training neural networks having a sparse topology.
Let the sensitivity of a network parameter be the variation of the loss function with respect to the variation of the parameter.
Parameters with low sensitivity, i.e. having little impact on the loss when perturbed, are shrunk and then pruned to sparsify the network.
Our method allows to train a network from scratch, i.e. without preliminary learning or rewinding. 
Experiments on multiple architectures and datasets show competitive compression ratios with minimal computational overhead.

\end{abstract}


\section{Introduction}

Artificial Neural Networks (ANNs) achieve state-of-the-art performance in several tasks at the price of complex topologies with millions of learnable parameters.
As an example, ResNet~\cite{he2016deep} includes tens of millions of parameters, soaring to hundreds of millions for VGG-Net~\cite{simonyan2014very}.
A large parameter count jeopardizes however the possibility to deploy a network over a memory-constrained (e.g., embedded, mobile) device, calling for leaner architectures with fewer parameters.

\noindent The complexity of ANNs can be reduced enforcing a \emph{sparse} network topology.
Namely, some connections between neurons can be \emph{pruned} by wiring the corresponding parameters to zero. Besides the reduction of parameters, some works also suggested other benefits coming from pruning ANNs, like improving the performance in transfer learning scenarios~\cite{liu2017sparse}.
Popular methods such as~\cite{han2015learning}, for example, introduce a \emph{regularization} term in the cost function with the goal to shrink to zero some parameters.
Next, a threshold operator pinpoints the shrunk parameters to zero, eventually enforcing the sought sparse topology.
However, such methods require that the topology to be pruned has been preliminarily pruned via standard gradient descent, which sums up to the total learning time.

\noindent This work contributes LOBSTER (\emph{LOss-Based SensiTivity rEgulaRization}), a method for learning sparse neural topologies.
In this context, let us define the \textit{sensitivity} of the parameter of an ANN as the derivative of the loss function with respect to the parameter.
Intuitively, low-sensitivity parameters have a negligible impact on the loss function when perturbed, and so are fit to be shrunk without compromising the network performance.
Practically, LOBSTER shrinks to zero parameters with low sensitivity with a regularize-and-prune approach, achieving a sparse network topology.
With respect to similar literature \cite{han2015deep, guo2016dynamic, gomez2019targeted}, LOBSTER does not require a preliminary training stage to learn the dense reference topology to prune. 
Moreover, differently to other sensitivity-based approaches, LOBSTER computes the sensitivity exploiting the already available gradient of the loss function, avoiding additional derivative computations~\cite{mozer1989skeletonization, tartaglione2018sensitivity}, or second-order derivatives~\cite{lecun1990optimal}.
Our experiments, performed over different network topologies and datasets, show that LOBSTER outperforms several competitors in multiple tasks.

\noindent The rest of this paper is organized as follows. In Sec.~\ref{sec:sota} we review the relevant literature concerning sparse neural architectures.
Next, in Sec.~\ref{sec:LDS} we describe our method for training a neural network such that its topology is sparse.
We provide a general overview on the technique in Sec.~\ref{sec:training}.
Then, in Sec.~\ref{sec:exp}  we experiment with our proposed training scheme over some deep ANNs on a number of different datasets.
Finally, Sec.~\ref{sec:conclusion} draws the conclusions while providing further directions for future research.

\section{Related Works}
\label{sec:sota}

It is well known that many ANNs, trained on some tasks, are typically over-parametrized ~\cite{mhaskar2016deep, brutzkus2017sgd}.
There are many ways to reduce the size of an ANN.
In this work we focus on the so-called \emph{pruning} problem: it consists in detecting and removing parameters from the ANN without excessively affecting its performance.
In a recent work~\cite{frankle2018lottery}, it has been observed that only a few parameters are actually updated during training: this suggests that all the others parameters can be removed from the learning process without affecting the performance.\attilio{Ma era proprio così la storia del paper ticket lottery ? MEssa così sembra che quel paper ha già stabilito quali sono i pesi da eliminare, e quindi il nostro lavoro a che serve ? Inoltre questo é un paper del 2020: se seguiamo ilc riterio cronologico, questo dobbiamo metterlo alla fine.}
Despite similar approaches were already taken years earlier~\cite{karnin1990simple}, their finding woke-up the research interest around such a topic.\attilio{Questa frase non é chiara né utile: dovremmo almeno dire cosa fa questo paper}
Lots of efforts are devoted towards making pruning mechanisms more efficient: for example, Wang~et~al. show that some sparsity is achievable pruning weights at the very beginning of the training process~\cite{wang2020pruning}, or Lee~et~al., with their ``SNIP'', are able to prune weights in a one-shot fashion~\cite{Lee2019SNIPSN}. However, these approaches achieve limited sparsity: iterative pruning-based strategy, when compared to one-shot or few-shot approaches, are able to achieve a higher sparsity~\cite{tartaglione2020pruning}. Despite the recent technology advances make this problem actual and relevant by the community towards the ANN architecture optimization, it deepens its roots in the past.\\
In Le~Cun~et~al.~\cite{lecun1990optimal}, the information from the second order derivative of the error function is leveraged to rank the the parameters of the trained model on a \emph{saliency} basis: this allows to select a trade-off between size of the network (in terms of number of parameters) and and performance.
In the same years, Mozer and Smolensky proposed \emph{skeletonization}, a technique to identify, on a trained model, the less relevant neurons, and to remove them~\cite{mozer1989skeletonization}. This is accomplished evaluating the global effect of removing a given neuron, evaluated as error function penalty from a pre-trained model.\\
\attilio{\cite{mozer1989skeletonization} é antecedente a \cite{lecun1990optimal}, lo metterei prima}
The recent technological advances let ANN models to be very large, and pose questions about the efficiency of pruning algorithms: the target of the technique is to achieve the highest \emph{sparsity} (ie. the maximum percentage of removed parameters) having minimal performance loss (accuracy loss from the ``un-pruned'' model). Towards this end, a number of different approaches to pruning exists.\\
Dropout-based approaches constitute another possibility to achieve sparsity. For example, \emph{Sparse~VD} relies on variational dropout to promote sparsity~\cite{molchanov2017variational}, providing also a Bayesian interpretation for Gaussian dropout. Another dropout-based approach is \emph{Targeted Dropout}~\cite{gomez2019targeted}: there, fine-tuning the ANN model is self-reinforcing its sparsity by stochastically dropping connections (or entire units).\\
Some approaches to introduce sparsity in ANNs attempt to rely on the optimal $\ell_0$ regularizer which, however, is a non-differentiable measure. A recent work~\cite{louizos2017learning} proposes a differentiable proxy measure to overcome this problem introducing, though, some relevant computational overhead. Having a similar overall approach, in another work, a regularizer based on group lasso whose task is to cluster filters in convolutional layers is proposed~\cite{wen2016structured}. However, such a technique is not directly generalizeable to the bulky fully-connected layers, where most of the complexity (as number of parameters) lies.\\
A sound approach towards pruning parameters consists in exploiting a $\ell_2$ regularizer in a shrink-and-prune framework. In particular, a standard $\ell_2$ regularization term is included in the minimized cost function (to penalize the magnitude of the parameters): all the parameters dropping below some threshold are pinpointed to zero, thus learning a sparser topology~\cite{han2015learning}.
Such approach is effective since regularization replaces unstable (ill-posed) problems with nearby and stable (well-posed) ones by introducing a prior on the parameters~\cite{Groetsch1993InverseProblems}.
However, as a drawback, this method requires a preliminary training to learn the threshold value; furthermore, all the parameters are blindly, equally-penalized by their $\ell_2$ norm: some parameters, which can introduce large error (if removed), might drop below the threshold because of the regularization term: this introduces sub-optimalities as well as instabilities in the pruning process. Guo et al. attempted to address this issue with their DNS~\cite{guo2016dynamic}: they proposed an algorithmic procedure to corrects eventual over-pruning by enabling the recovery of severed connections.
Moving to sparsification methods not based on pruning, \textit{Soft Weight Sharing} (SWS)~\cite{ullrich2017soft} shares redundant parameters among layers, resulting in fewer parameters to be stored. Approaches based on knowledge distillation, like \textit{Few Samples Knowledge Distillation} (FSKD)~\cite{li2020few}, are also an alternative to reduce the size of a model: it is possible to successfully train a small student network from a larger teacher, which has been directly trained on the task. Quantization can also be considered for pruning: Yang~et~al., for example, considered the problem of ternarizing and prune a pre-trained deep model~\cite{yang2020harmonious}. Other recent approaches mainly focus on the pruning of convolutional layers either leveraging on the artificial bee colony optimization algorithm (dubbed as ABCPruner)~\cite{lin2020channel} or using a small set of input to evaluate a saliency score and construct a sampling distribution~\cite{liebenwein2020provable}.\\
In another recent work~\cite{tartaglione2018sensitivity}, it was proposed to measure how much the network output changes for small perturbations of some parameters, and to iteratively penalize just those which generate little or no performance loss. However, such method requires the network to be already trained so to measure the variation of the network output when a parameter is perturbed, increasing the overall learning time.\\
In this work, we overcome the basic limitation of pre-training the network, introducing the concept of \emph{loss-based sensitivity}: it only penalizes the parameters whose small perturbation introduces little or no performance loss at training time. 

\section{Proposed Regularization}
\label{sec:LDS}

\attilio{
Proposta: 
In this section we first express the sensitivity of a network parameter as the variation the loss function as a function of a perturbation applied to the parameter.
Then, we propose a parameter update rule that includes a regularization term accounting for each parameter sensitivity to driving towards zero small sensitivity parameters.
}

\subsection{Loss-based Sensitivity}
\label{sec::outsens}

ANNs are typically trained via gradient descent based optimization, i.e. minimizing the loss function \attilio{$L(\textbf{w} ,\textbf{y})$}.
Methods based on mini-batches of samples have gained popularity as they allow better generalization than stochastic learning while they are memory and time efficient.
In such a framework, a network parameter $w_i$ is updated towards the averaged direction which minimizes the averaged loss for the minibatch, i.e. using the well known stochastic gradient descent or its variations.
If the gradient magnitude is close to zero, then the parameter is not modified. \attilio{ messa così, sembra che ci sia una threshold nel SGD, per cui se il modulo del gradiente e sotto questa threshold il parametro non viene updated. Immagino volessi dire che é proporzionale ?}
Our ultimate goal is to assess to which extent a variation of the value of $w_i$ would affect the error on the network output $\mathbf{y}$.
\attilio{The underlying rationale is that parameters not affecting the network output whatever their value, can be safely hardwired to zero, i.e. pruned away.}
We make a first attempt towards this end introducing a small perturbation $\Delta w_i$ over $w_i$ and measuring the variation of $\mathbf{y}$ as

\begin{equation}
    \label{eq:sensitivity_old}
    \Delta \mathbf{y} = \sum_k \left| \Delta y_k \right| \approx \Delta w_i \sum_k \left| \frac{\partial y_k}{\partial w_i} \right|.
\end{equation}
\noindent
Unfortunately, the evaluation of \eqref{eq:sensitivity_old} is specific and restricted to the neighborhood of the network output.
\attilio{Rileggendo la frase sopra, mi chiedevo: forse volevi dire é ristretto ad un intorno \textit{del valore attuale} di y come funzioen del particolare wi? }
We would like to directly evaluate the error of the output of the ANN model over the learned data. \attilio{Questa ultima frase non é chiara}

\begin{table}
    \center
    \begin{tabular}{cccc}
        \toprule
        $P\left(\frac{\partial L}{\partial w_{i}}\right)$ & $\sign\left(\frac{\partial L}{\partial w_{i}}\right )$ & $\sign\left(w \right )$ & $\frac{\tilde{\eta}}{\eta}$ \\
        \midrule
        0       &any    &any    &$1$\\
        1       &+      &+      &$\leq 1$\\
        1       &+      &-      &$\geq 1$\\
        1       &-      &+      &$\geq 1$\\
        1       &-      &-      &$\leq 1$\\
        \bottomrule
    \end{tabular}
    \caption{Behavior of $\tilde{\eta}$ compared to $\eta$ ($\eta > 0$).}
    \label{tab:eta}
\end{table}

Towards this end, we estimate the error on the network output caused by the perturbation on $w_i$ \attilio{by measuring the variation of the loss function} as:

\begin{equation}
    \label{eq:sensitivity_old2}
    \Delta L \approx \Delta w_i \left | \frac{\partial L}{\partial \mathbf{y}} \cdot \frac{\partial \mathbf{y}}{\partial w_i} \right| = \Delta w_i \left | \frac{\partial L}{\partial w_i}\right|.
\end{equation}
\noindent
The use of \eqref{eq:sensitivity_old2} in place of \eqref{eq:sensitivity_old} shifts the focus from the output to the error of the network.
The latter is a more accurate information in order to evaluate the real effect of the perturbation of a given parameter $w_i$ \attilio{In che senso é più accurata ? Che non dipende solo dal particolare valore assunto da wi in quel momento ?}.
Let us define the \textit{sensitivity} $S$ for a given parameter $w_i$ as

\begin{equation}
    \label{eq:S}
    S(L, w_i) = \left | \frac{\partial L}{\partial w_i}\right|.
\end{equation}
\noindent
Large $S$ values indicate large variations of the loss function \attilio{\textit{even} ? } for small perturbations of $w_i$.

Given the above sensitivity definition, we can promote sparse topologies by pruning parameters with both low sensitivity  $S$ (i.e., in a flat region of the loss function gradient, where a small perturbation of the parameter has a negligible effect on the loss) and low magnitude, keeping unmodified those with large $S$ \attilio{Questa ultima parte dopo la virgola vuol solo dire che i parametri con S grande non li agggiorniamo nemmeno o che semplicemente non li pruniamo ?}.
Towards this end, we propose the following parameter update rule to promote sparsity:

\begin{align}
    w_{i}^{t+1} := w_{i}^t &- \eta \frac{\partial L}{\partial w_{i}^t}+\nonumber \\
    &- \lambda w_{i}^t \left[1 - S(L, w_{i}^t)\right] P\left[S(L, w_{i}^t)\right],
    \label{eq:oleupdateRule}
\end{align}
where
\begin{equation}
    P(x) = \Theta\left[1 - |x| \right],
\end{equation}
$\Theta(\cdot)$ is the one-step function and $\eta, \lambda$ two positive hyper-parameters. 
 
\subsection{Update Rule}
\attilio{Qui inizia la sezione \textit{Update rule} ma in realta noi abbiamo definito una prima update rule poco sopra e il lettore si chiede \textit{Ma quella di eq 4 non era gia una update rule?}. Proporrei di fare iniziare questa subsection subito prima della frase \textit{Given the above sensitivity definition, we ...} e di togliere \textit{Previously we have introduced a measure for the sensitivity which can be used also at training time.}}
\label{sec:ur}
Previously we have introduced a measure for the sensitivity which can be used also at training time. In particular, plugging \eqref{eq:S} in \eqref{eq:oleupdateRule} we can rewrite the update rule as:

\begin{equation}
    w_{i}^{t+1} = w_{i}^t - \eta \frac{\partial L}{\partial w_{i}^{t}} - \lambda \Gamma\left(L, w_{i}^{t}\right) \left[1 - \left|\frac{\partial L}{\partial w_{i}^{t}}\right | \right],
    \label{eq:newuprule}
\end{equation}
where
\begin{equation}
    \Gamma\left(y, x \right) = x \cdot P\left( \frac{\partial y}{\partial x} \right).
\end{equation}
After some algebraic manipulations, we can rewrite \eqref{eq:newuprule} as
\begin{align}
    w_{i}^{t+1} &= w_{i}^{t} - \lambda\Gamma\left(L, w_{i}^{t} \right)+\nonumber \\
    &- \frac{\partial L}{\partial w_{i}^{t}}\left[ \eta - \sign\left(\frac{\partial L}{\partial w_{i}^{t}}\right )\lambda\Gamma\left(L, w_{i}^{t} \right)  \right].
    \label{eq:newuprule2}
\end{align}

\begin{figure}
    \begin{subfigure}{0.45\columnwidth}
            \centering
            \includegraphics[width=\textwidth]{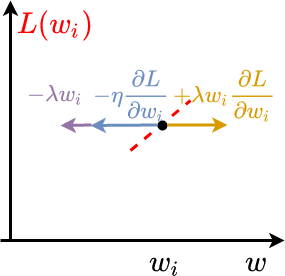}
            \caption{~}
            \label{fig::casea}
    \end{subfigure}
    \hfill
    \begin{subfigure}{0.45\columnwidth}
            \centering
            \includegraphics[width=\textwidth]{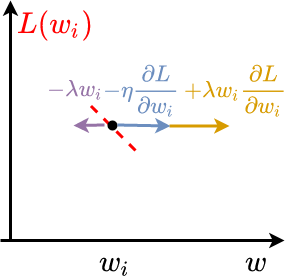}
            \caption{~}
            \label{fig::caseb}
    \end{subfigure}
    
    \vskip\baselineskip
    
    \begin{subfigure}{0.45\columnwidth}
            \centering
            \includegraphics[width=\textwidth]{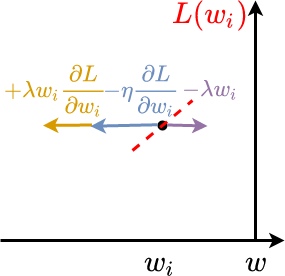}
            \caption{~}
            \label{fig::casec}
    \end{subfigure}%
    \hfill
    \begin{subfigure}{0.45\columnwidth}
            \centering
            \includegraphics[width=\textwidth]{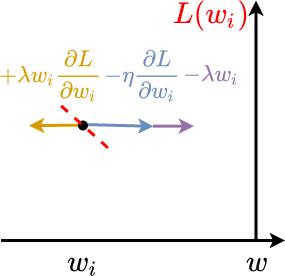}
            \caption{~}
            \label{fig::cased}
    \end{subfigure}
    \caption{Update rule effect on the parameters. The red dashed line is the tangent to the loss function in the black dot, in blue the standard SGD contribution, in purple the weight decay while in orange the LOBSTER contribution. Here we assume $P(L, w_{i})=1$.}
    \label{fig::effectLR}
\end{figure}
\noindent
In~\eqref{eq:newuprule2}, we observe two different components of the proposed regularization term:
\begin{itemize}
    \item a weight decay-like term $\Gamma\left(L, w_{i} \right)$ which is enabled/disabled by the magnitude of the gradient on the parameter;
    \item a correction term for the learning rate. In particular, the full learning process follows an \emph{equivalent} learning rate
    \begin{equation}
        \tilde{\eta} = \eta - \sign\left(\frac{\partial L}{\partial w_{i}}\right )\lambda\Gamma\left(L, w_{i} \right).
    \end{equation}
\end{itemize}
Let us analyze the corrections in the learning rate.
If $\left|\frac{\partial L}{\partial w_{i}}\right|\geq 1$ ($w_i$ has large sensitivity), it follows that $P\left( \frac{\partial L}{\partial w_{i}}\right)=0$ and $\Gamma\left(L, w_{i} \right)=0$ and the dominant contribution comes from the gradient.
In this case our update rule reduces to the classical GD:
\begin{equation}
    w_{i}^{t+1} = w_{i}^{t} - \eta \frac{\partial L}{\partial w_{i}^{t}}.
\end{equation}
\noindent
When we consider less sensitive $w_{i}$ with $\left|\frac{\partial L}{\partial w_{i}}\right|< 1$, we get $\Gamma \left(L, w_{i} \right) = w_{i}$ (weight decay term) and we can distinguish two sub-cases for the learning rate:
\begin{itemize}
    \item if $\sign\left(\frac{\partial L}{\partial w_{i}}\right )= \sign\left(w_i \right )$, then $\tilde{\eta} \leq \eta$ (Fig.~\ref{fig::casea} and Fig.~\ref{fig::cased}),
    \item if $\sign\left(\frac{\partial L}{\partial w_{i}}\right )\neq \sign\left(w_i \right )$, then $\tilde{\eta} \geq \eta$ (Fig.~\ref{fig::caseb} and Fig.~\ref{fig::casec}).
\end{itemize}
A schematics of all these cases can be found in Table~\ref{tab:eta} and the representation of the possible effects are shown in Fig.~\ref{fig::effectLR}. The contribution coming from $\Gamma\left(L, w_{i} \right)$ aims at minimizing the parameter magnitude, disregarding the loss minimization. If the loss minimization tends to minimize the magnitude as well, then the equivalent learning rate is reduced. On the contrary, when the gradient descent tends to increase the magnitude, the learning rate is increased, to compensate the contribution coming from $\Gamma\left(L, w_{i} \right)$. This mechanism allows us to succeed in the learning task while introducing sparsity.\\
In the next section we are going to detail the overall training strategy, which cascades a \emph{learning} and a \emph{pruning} stage. 

\section{Training Procedure}
\label{sec:training}

\begin{figure*}
    \captionsetup[subfigure]{position=b}
    \subcaptionbox{~ \label{fig::FG}}{\includegraphics[width=.33\textwidth]{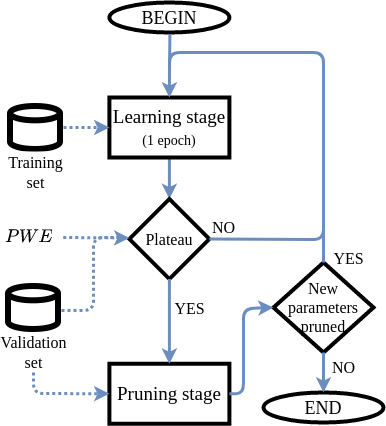}}
    \hfill
    \subcaptionbox{~ \label{fig::pruning}}{\includegraphics[width=.63\linewidth]{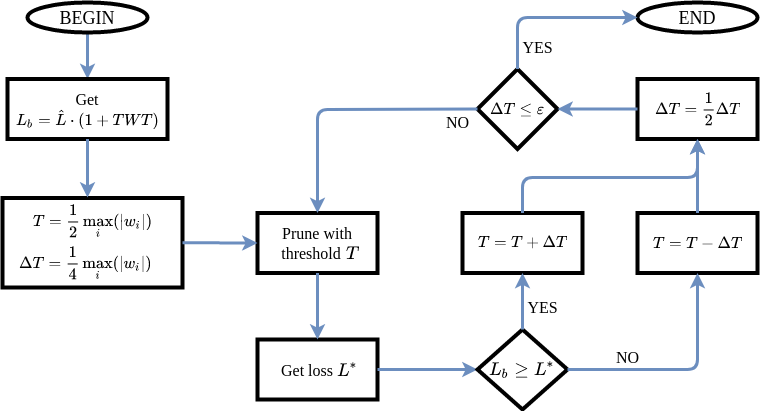}}
    \caption{The complete training procedure of LOBSTER~(a) and detail of the pruning stage~(b).}
    \label{fig::training}
\end{figure*}

This section describes a procedure to train a sparse neural network $\mathcal{N}$ leveraging the sensitivity-based rule above to update the network parameters.
We assume that the parameters have been randomly initialized, albeit the procedure holds also if the network has been pre-trained.
The procedure is illustrated in Fig.~\ref{fig::FG} and iterates over two stages as follows.

\subsection{Learning Stage}

During the learning stage, the ANN is iteratively trained according to the update rule~\eqref{eq:oleupdateRule} on some training set.
Let $e$ indicate the current learning stage iteration (i.e., epoch) and $\mathcal{N}^{e}$ represent the network (i.e., the set of learnable parameters) at the end of the $e$-th iteration. Also let $L^{e}$ be the loss measured on some validation set at the end of the $e$-th iteration and $\widehat{L}$ be the best (lowest) loss measured so far on $\widehat{\mathcal{N}}$ (network with lowest validation loss so far).
As initial condition, we assume, $\widehat{\mathcal{N}} = \mathcal{N}^0$.
If $L^{e} < \widehat{L}$, the reference to the  best network is updated as $\widehat{\mathcal{N}} = \mathcal{N}^{e}$, $\widehat{L} = L^{e}$.
We iterate again the learning stage $\mathcal{N}$ until the best validation loss $L^{e}$ has not decreased for $PWE$ iterations of the learning stage in a row (we say the regularizer has reached a performance \emph{plateau}). At such point, we move to the pruning stage.\\
We provide $\widehat{\mathcal{N}}$ as input for the pruning stage, where a number of parameters have been shrunk towards zero by our sensitivity-based regularizer.

\subsection{Pruning Stage}
\label{sec:pruning}

In a nutshell, during the pruning stage parameters with magnitude below a threshold value $T$ are pinpointed to zero, eventually sparsifying the network topology as shown in Fig.~\ref{fig::pruning}.
Namely, we look for the largest $T$ that worsens the classification loss $L_b$ at most by a relative quantity $TWT$:
\begin{equation}
    L^b =  \left(1+TWT \right) \widehat{L},
\end{equation}
\noindent
where $L^b$ is called \emph{loss boundary}.
$T$ is found using the bisection method, initializing $T$ with the average magnitude of the non-null parameters in the network. Then, we apply the threshold $T$ to $\widehat{\mathcal{N}}$ obtaining the pruned network $\mathcal{N}^*$ with its loss $L^*$ on the validation set.
At the next pruning iteration, we update $T$ as follows:
\begin{itemize}
    \item if $L_b \geq L^*$ the network tolerates that more parameters be pruned, so $T$ is increased;
    \item if $L_b < L^*$ then too many parameters have been pruned and we need to restore some: we decrease $T$.
\end{itemize}
The pruning stage ends when $L_b = L^*$ and we observe that $L_b < L^*$ for any new threshold $T+\varepsilon~\forall~\varepsilon > 0$.
Once $T$ is found, all the parameters whose magnitude is below $T$ are pinpointed to zero, i.e. they are pruned for good.
If at leas one parameter has been pruned during the last iteration of the pruning stage, a new iteration of the regularization stage follows; otherwise, the procedure ends returning the trained, sparse network.

\section{Results}
\label{sec:exp}

\begin{figure*}
    \captionsetup[subfigure]{position=b}
    \subcaptionbox{~ \label{fig::ln300}}{\includegraphics[width=.45\textwidth]{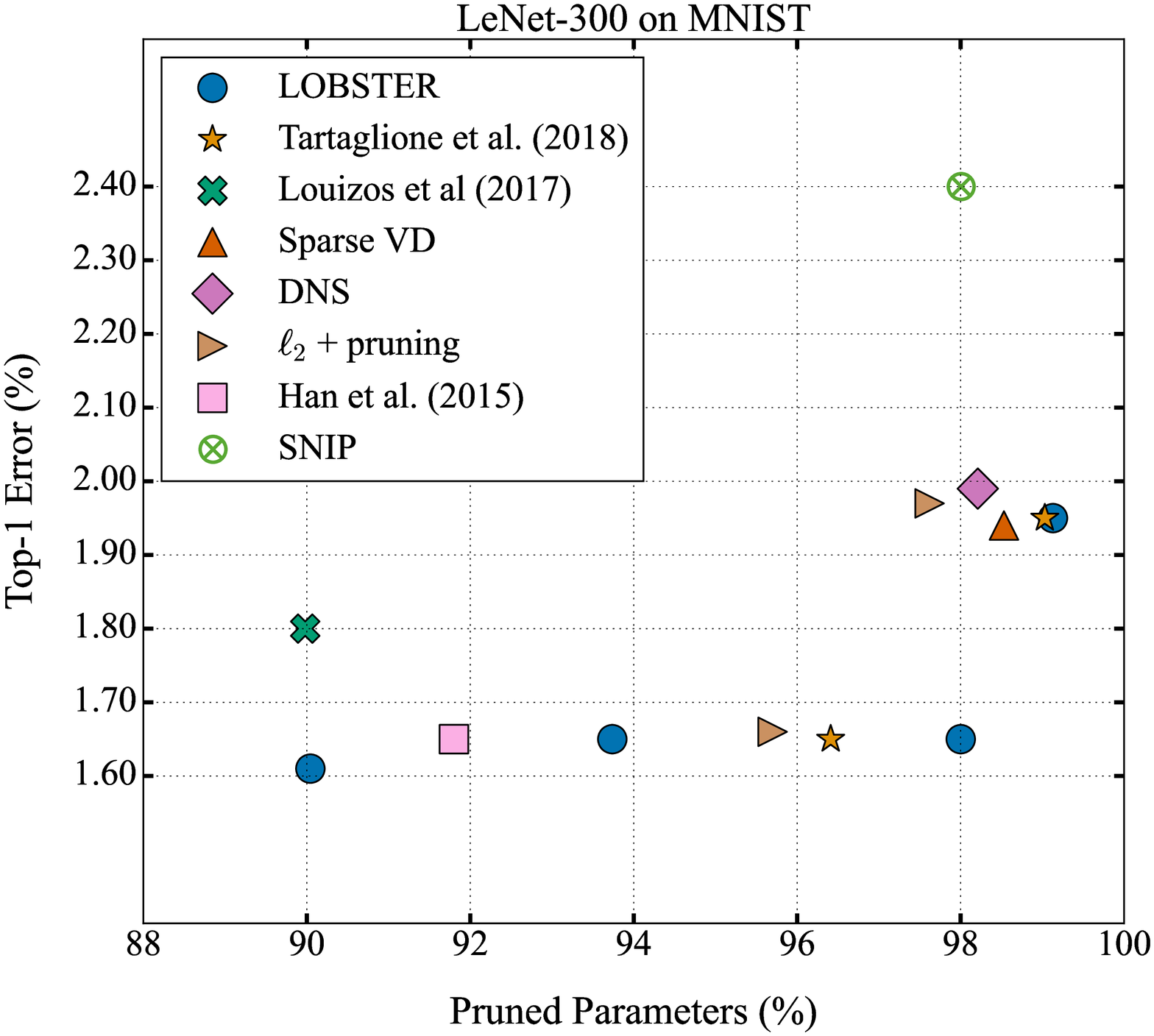}}
    \hfill
    \subcaptionbox{~ \label{fig::ln5_m}}{\includegraphics[width=.45\linewidth]{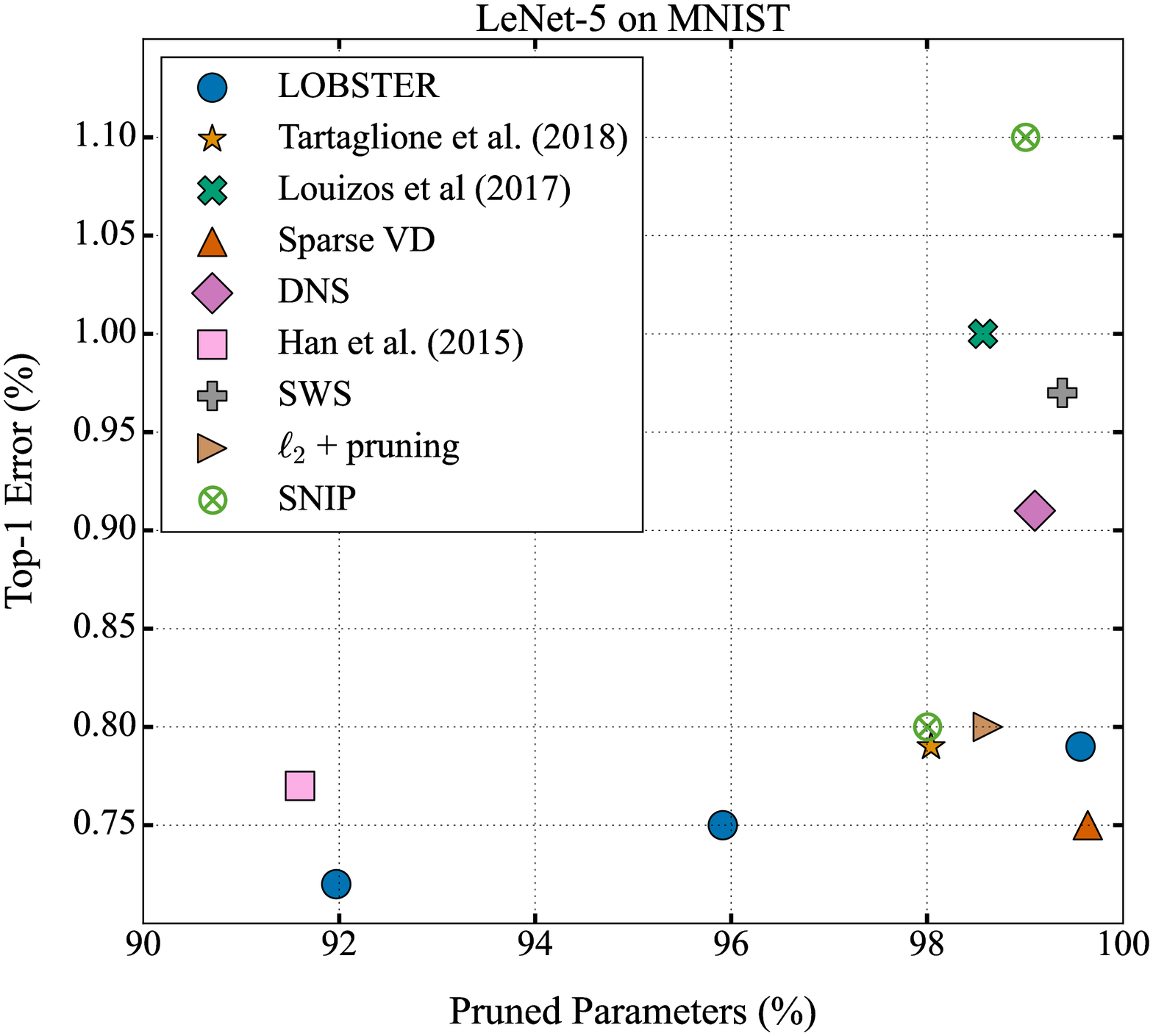}}
    
    \subcaptionbox{~ \label{fig::ln5_f}}{\includegraphics[width=.45\textwidth]{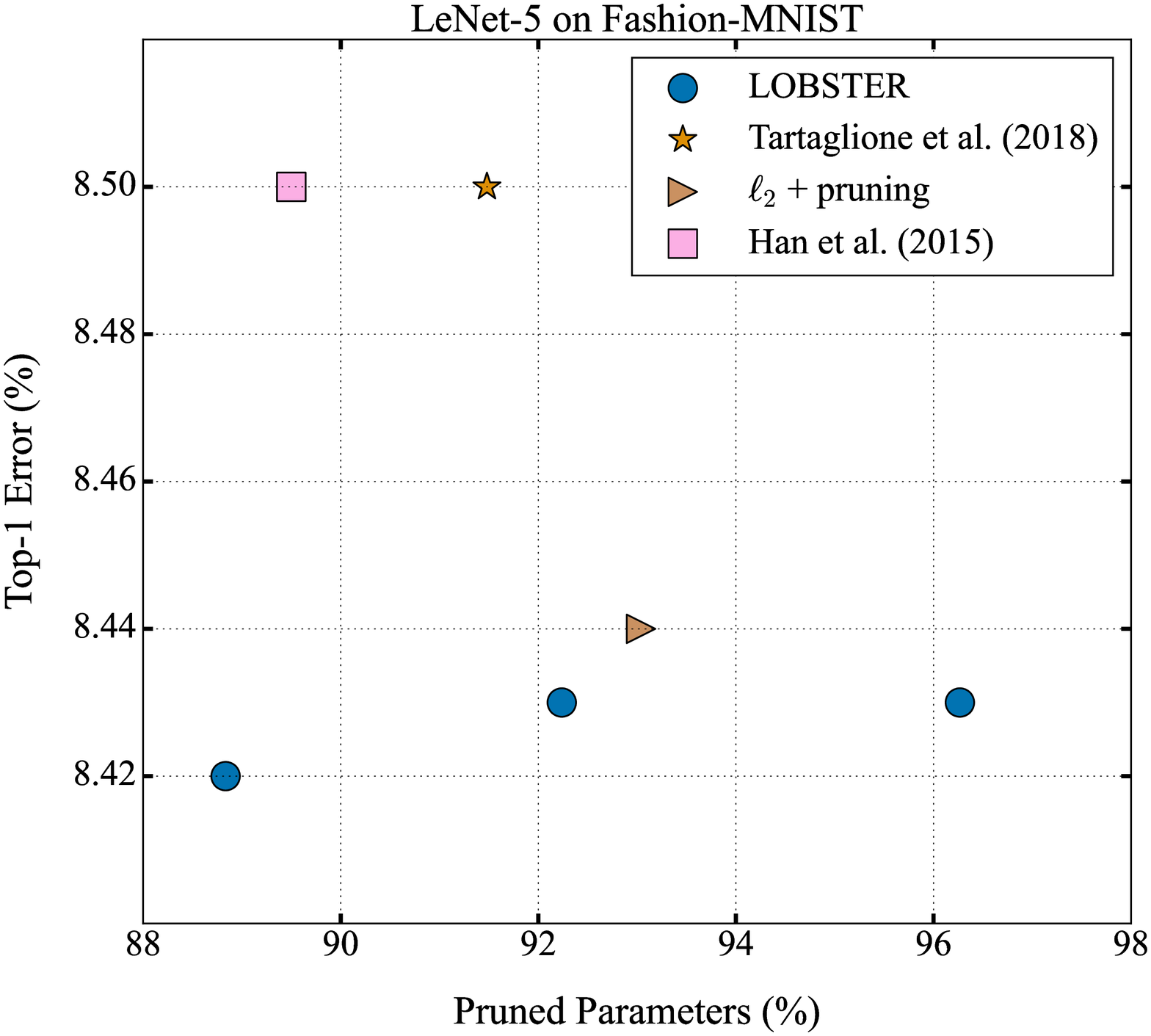}}
    \hfill
    \subcaptionbox{~ \label{fig::res32}}{\includegraphics[width=.45\linewidth]{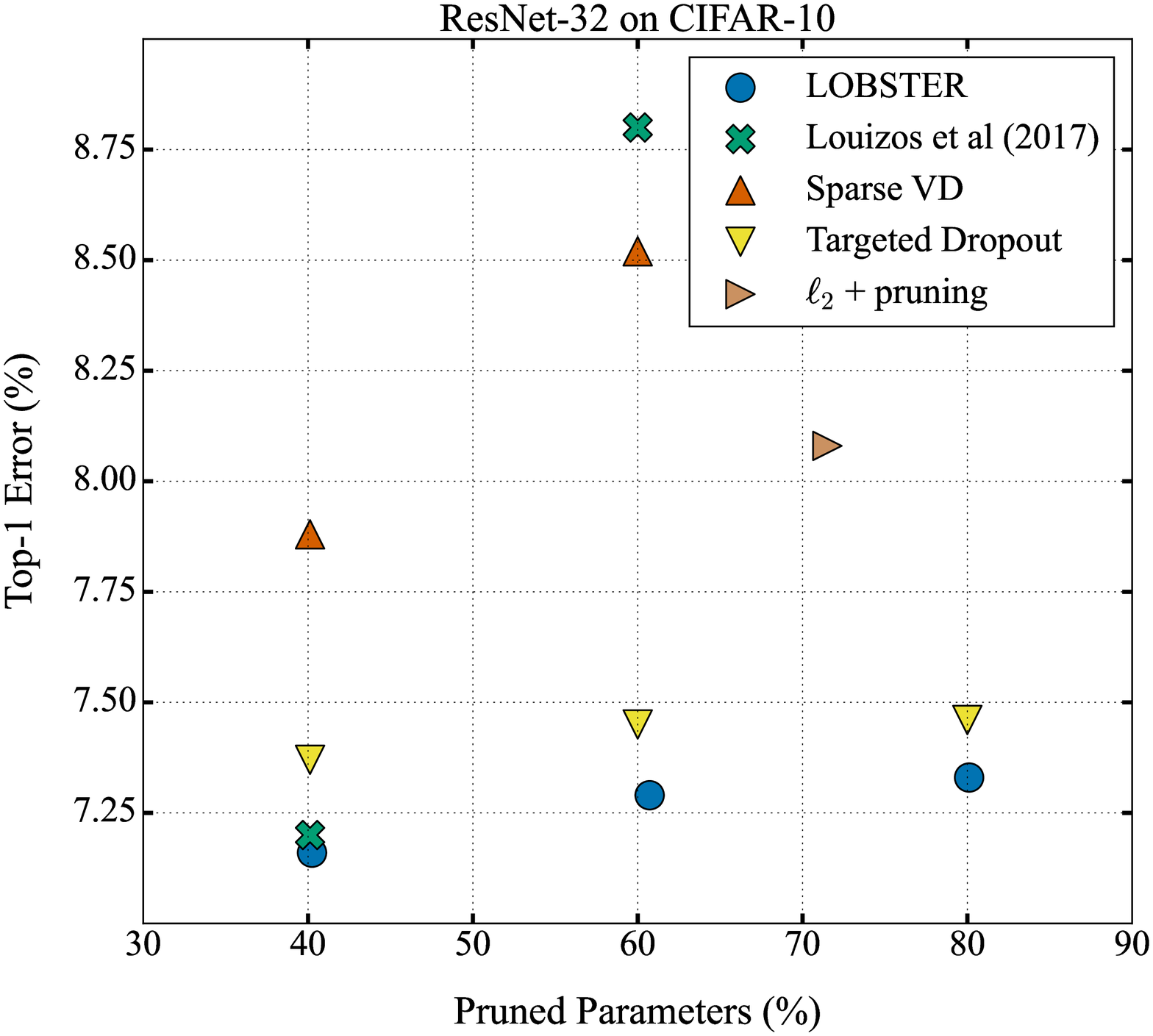}}
    
    \subcaptionbox{~ \label{fig::res18}}{\includegraphics[width=.45\textwidth]{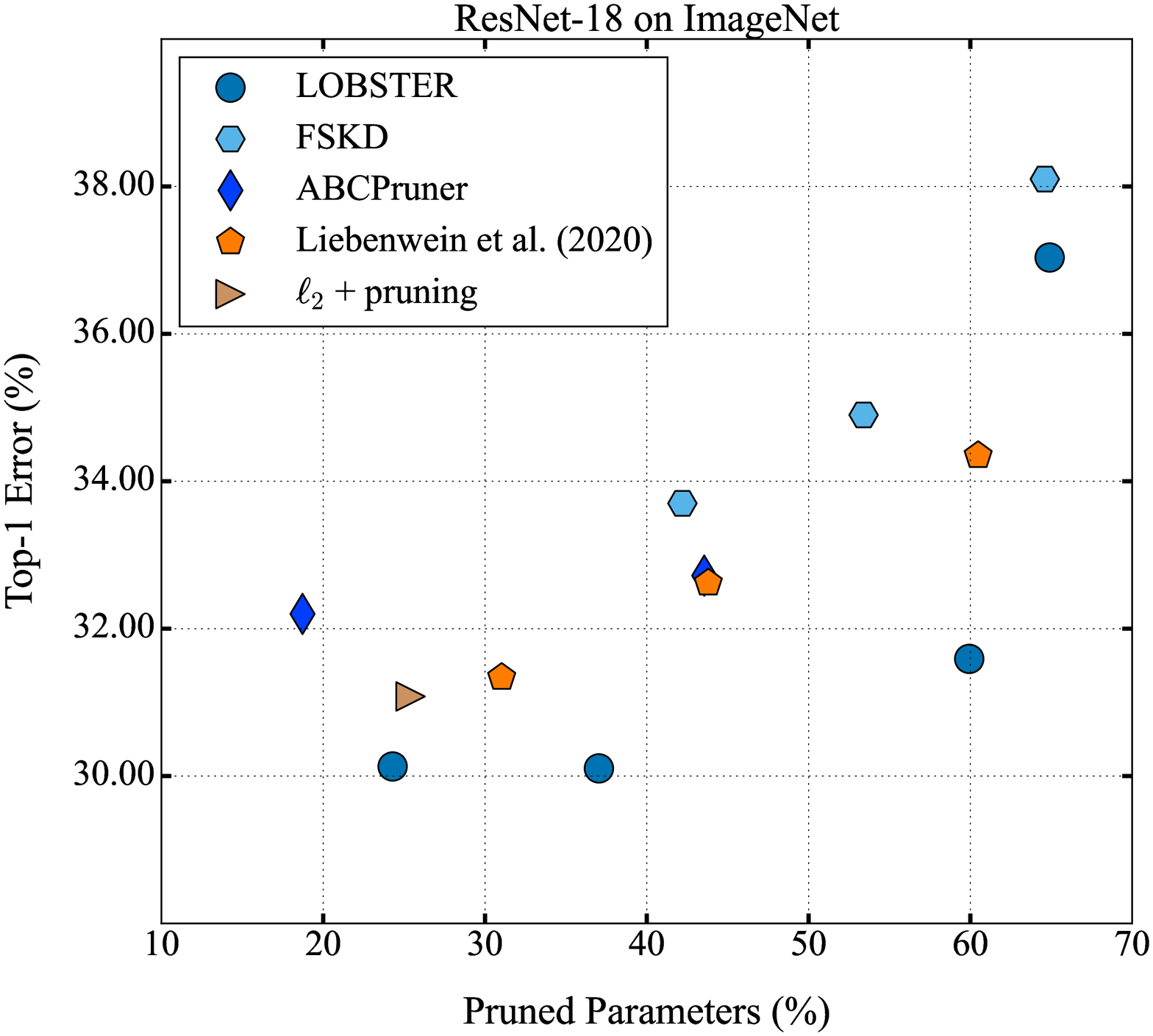}}
    \hfill
    \subcaptionbox{~ \label{fig::res101}}{\includegraphics[width=.45\linewidth]{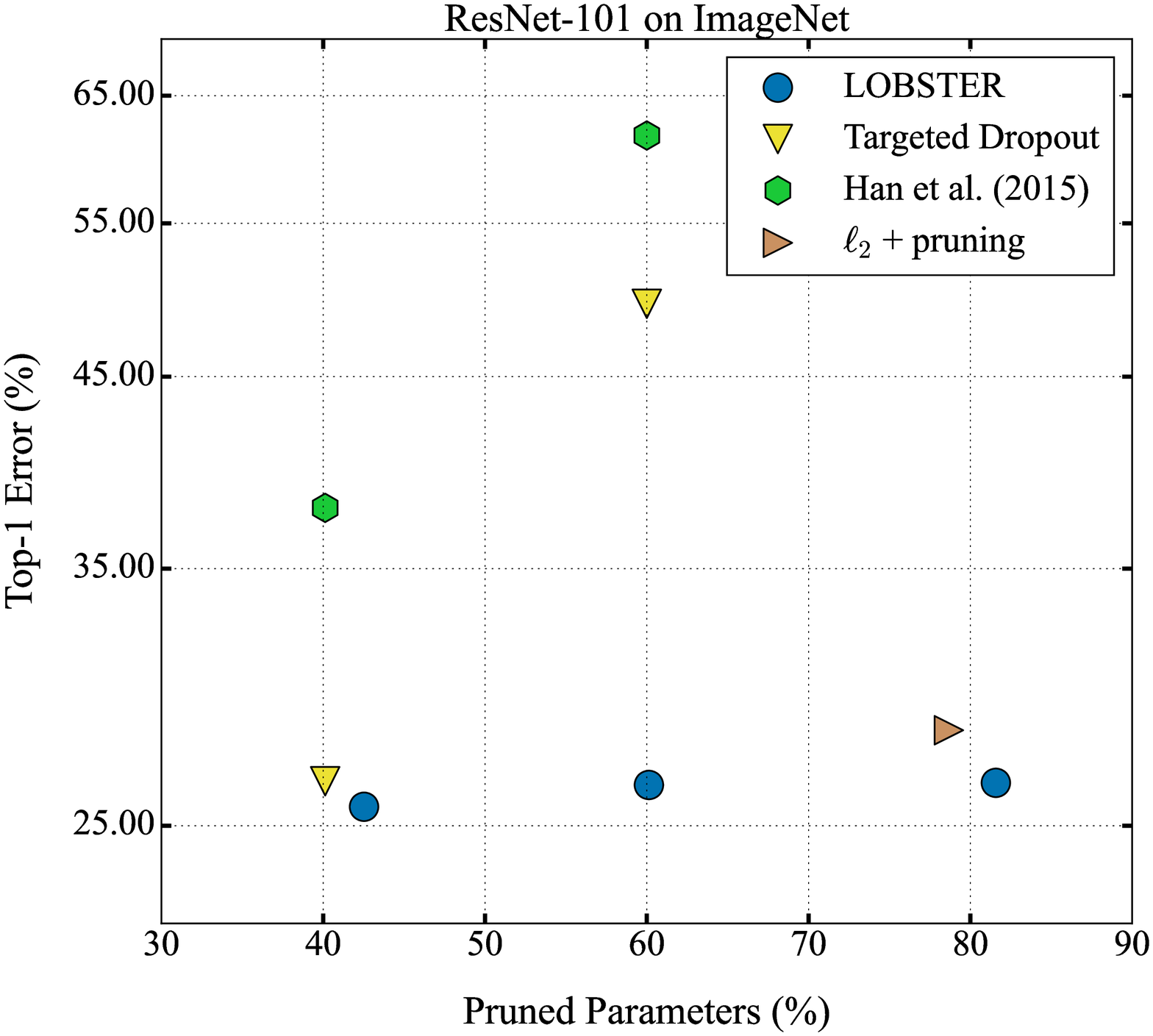}}
    \caption{Performance (Top-1 error) vs ratio of pruned parameters for LOBSTER and other state of the art methods over different architectures and datasets.}
    \label{fig::globalresults}
\end{figure*}

In this section we experimentally evaluate LOBSTER over multiple architectures and datasets commonly used as benchmark in the literature:

\begin{itemize}
    \item LeNet-300 on MNIST (Fig.~\ref{fig::ln300}),
    \item LeNet-5 on MNIST (Fig.~\ref{fig::ln5_m}),
    \item LeNet-5 on Fashion-MNIST (Fig.~\ref{fig::ln5_f}),
    \item ResNet-32 on CIFAR-10 (Fig.~\ref{fig::res32}),
    \item ResNet-18 on ImageNet (Fig.~\ref{fig::res18}),
    \item ResNet-101 on ImageNet (Fig.~\ref{fig::res101}).
\end{itemize}

\noindent We compare with other state-of-the-art approaches introduced in Sec.~\ref{sec:sota} wherever numbers are publicly available. Besides these, we also perform an ablation study with a $\ell_2$-based regularizer and our proposed pruning strategy (as discussed in Sec.~\ref{sec:pruning}). Performance is measured as the achieved model sparsity versus classification error (Top-1 or Top-5 error). The network sparsity is defined here ad the percentage of pruned parameters in the ANN model. 
Our algorithms are implemented in Python, using PyTorch~1.2 and simulations are run over an RTX2080~TI NVIDIA GPU. All the hyper-parameters have been tuned via grid-search.
The validation set size for all the experiments is 5k large. \footnote{The source code is provided in the supplementary materials and will be made publicly available upon acceptance of the article.}
For all datasets, the learning and pruning stages take place on a random split of the training set, whereas the numbers reported below are related to the test set.

\subsection{LeNet-300 on MNIST}


As a first experiment, we train a sparse LeNet-300~\cite{LeCunPrIEEE1998DocumentRecognition} architecture, which consists of three fully-connected layers with 300, 100 and 10 neurons respectively. We trained the network on the MNIST dataset, made of 60k training images and 10k test gray-scale 28$\times$28 pixels large images, depicting handwritten digits. Starting from a randomly initialized network, we trained LeNet-300 via SGD with learning rate $\eta = 0.1$, $\lambda=10^{-4}$, $PWE=20$ epochs and $TWT=0.05$.\\
\noindent The related literature reports several compression results that can be clustered in two groups corresponding to classification error rates of about $1.65\%$ and $1.95\%$, respectively.
Fig.~\ref{fig::ln300} provides results for the proposed procedure. Our method reaches higher sparsity than the the approaches found in literature.
This is particularly noticeable around $1.65\%$ classification error (low left in Fig.~\ref{fig::ln300}), where we achieve almost twice the sparsity of the second best method. 
LOBSTER also achieves the highest sparsity for the higher error range (right side of the graph), gaining especially in regards to the number of parameters removed from the first fully-connected layer (the largest, consisting of 235k parameters), in which we observe just the $0.59\%$ of the parameters survives.

\begin{table*}
	\renewcommand{\arraystretch}{1.3}
    \vskip 0.1in
	\centering
		\begin{tabular}{c c   c c c c c c}
			\toprule
			\multirow{2}{*}{Dataset} & \multirow{2}{*}{Architecture} &\multicolumn{3}{c}{$\ell_2$+pruning}&\multicolumn{3}{c}{LOBSTER}\\
			&& Top-1 (\%) & Sparsity (\%) & FLOPs & Top-1 (\%) & Sparsity (\%) & FLOPs \\
			\midrule
			\multirow{2}{*}{MNIST}   
			        & LeNet-300     &     1.97      &     97.62      &22.31k     &      1.95     &      99.13     &10.63k\\
			        & LeNet-5       &     0.80      &      98.62     &589.75k    &     0.79      &      99.57     &207.38k\\
			\midrule
			F-MNIST &  LeNet-5      &      8.44     &      93.04     &1628.39k   &      8.43     &      96.27     &643.22k\\
			\midrule
			CIFAR-10& ResNet-32     &    8.08       &     71.51      &44.29M     &      7.33      &     80.11     &32.90M\\
			\midrule
			\multirow{2}{*}{ImageNet}
			        &ResNet-18      &       31.08    &      25.40     &2.85G    &      30.10     &      37.04     &2.57G     \\
			        &ResNet-101     &      28.33     &      78.67     &3.44G      &       26.44    &      81.58     &3.00G\\
			
			\bottomrule
	\end{tabular}
    \vskip 0.1in
    \caption{Comparison between LOBSTER and with $\ell_2$+pruning as in Fig.~\ref{fig::globalresults} (only best sparsity results are reported).}
    \label{tab:FLOPs}
\end{table*}

\subsection{LeNet-5 on MNIST and Fashion-MNIST}

Next, we experiment on the caffe version of the LeNet-5 architecture, consisting in two convolutional and two fully-connected layers. Again, we use a randomly-initialized network, trained via SGD with learning rate $\eta=0.1$, $\lambda=10^{-4}$, $PWE=20$ epochs and $TWT=0.05$. The results are shown in Fig.~\ref{fig::ln5_m}.
Even with a convolutional architecture, we obtain a competitively small network with a sparsity of 99.57\%. 
At higher compression rates, Sparse~VD slightly outperforms all other methods in the LeNet5-MNIST experiment.
We observe that LOBSTER, in this experiment, sparsifies the first convolutional layer ($22\%$ sparsity) more than Sparse~VD solution ($33\%$). In particular, LOBSTER prunes $14$ filters out of the $20$ original in the first layer (or in other words, just $6$ filters survive, and contain all the un-pruned parameters).
We hypothesize that, in the case of Sparse~VD and for this particular dataset, extracting a larger variety of features at the first convolutional layer, both eases the classification task (hence the lower Top-1 error) and allows to drop more parameters in the next layers (a slightly improved sparsity). However, since we are above $99\%$ of sparsity, the difference between the two techniques is minimal.\\
To scale-up the difficulty of the training task, we experimented on the classification of the Fashion-MNIST dataset~\cite{xiao2017fashionmnist}, using again LeNet5. This dataset has the same size and image format of the MNIST dataset, yet it contains images of clothing items, resulting in a non-sparse distribution of the pixel intensity value. Since the images are not as sparse, such dataset is notoriously harder to classify than MNIST. For this experiment, we trained the network from scratch using SGD with $\eta=0.1$, $\lambda=10^{-4}$, $PWE=20$ epochs and $TWT=0.1$. The results are shown in Fig.~\ref{fig::ln5_f}.\\
F-MNIST is an inherently more challenging dataset than MNIST, so the achievable sparsity is lower. Nevertheless, the proposed method still reaches higher sparsity than other approaches, removing an higher percentage of parameters, especially in the fully connected layers, while maintaining good generalization. In this case, we observe that the first layer is the least sparsified: this is an effect of the higher complexity of the classification task, which requires more features to be extracted.

\subsection{ResNet-32 on CIFAR-10}
\label{sec:res32exp}


To evaluate how our method scales to deeper, modern architectures, we applied it on a PyTorch implementation of the ResNet-32 network~\cite{kaiming2015resnet} that classifies the CIFAR-10 dataset.\footnote{\url{https://github.com/akamaster/pytorch_resnet_cifar10}} This dataset consists of 60k 32$\times$32 RGB images divided in 10 classes (50k training images and 10k test images). We trained the network using SGD with momentum $\beta=0.9$, $\lambda=10^{-6}$, $PWE=10$ and $TWT=0$. The full training is performed for 11k epochs.
Our method performs well on this task and outperforms other state-of-the-art techniques. Furthermore, LOBSTER improves the network generalization ability reducing the baseline Top-1 error from $7.37\%$ to $7.33\%$ of the sparsified network while removing $80.11\%$ of the parameters.
This effect is most likely due to the LOBSTER technique itself, which self-tunes the regularization on the parameters as explained in Sec.~\ref{sec:ur}.

\subsection{ResNet on ImageNet}

Finally, we further scale-up both the output and the complexity of the classification problem testing the proposed method on network over the well-known ImageNet dataset (ILSVRC-2012), composed of more than 1.2 million train images
, for a total of 1k classes. For this test we used SGD with momentum $\beta=0.9$, $\lambda=10^{-6}$ and $TWT=0$. 
The full training lasts 95 epochs. Due to time constraints, we decided to use the pre-trained network offered by the torchvision library.\footnote{\url{https://pytorch.org/docs/stable/torchvision/models.html}} 
Fig.~\ref{fig::res18} shows the results for ResNet-18 while Fig.\ref{fig::res101} shows the results for ResNet-101. 
Even in this scenario, LOBSTER proves to be particularly efficient: we are able to remove, with no performance loss, $37.04\%$ of the parameters from ResNet-18 and $81.58\%$ from ResNet-101.

\subsection{Ablation study}

As a final ablation study, we replace our sensitivity-based regularizer with a simpler $\ell_2$ regularizer in our leraning scheme in Fig.~\ref{fig::training}.
Such scheme ``$\ell_2$+pruning'' uniformly applies an $\ell_2$ penalty to all the parameters regardless their contribution to the loss.
This scheme is comparable with~\cite{han2015learning}, yet enhanced with the same pruning strategy with adaptive thresholding shown in Fig.~\ref{fig::pruning}.
A comparison between LOBSTER and $\ell_2$+pruning is reported in Table~\ref{tab:FLOPs}.\\
In all the experiments we observe that dropping the sensitivity based regularizer impairs the performance. 
This experiment verifies the role of the sensitivity-based regularization in the performance of our scheme.
Finally, Table~\ref{tab:FLOPs} also reports the corresponding inference complexity in FLOPs.
For the same or lower Top-1 error LOBSTER yelds benefits as fewer operations at inference time and suggesting the presence of some structure in the sparsity achieved by LOBSTER.

\section{Conclusion}
\label{sec:conclusion}

We presented LOBSTER, a regularization method suitable to train neural networks with a sparse topology without a preliminary training.
Differently from $\ell_2$ regularization, LOBSTER is aware of the global contribution of the parameter on the loss function and self-tunes the regularization effect on the parameter depending on factors like the ANN architecture or the training problem itself (in other words, the dataset). Moreover, tuning its hyper-parameters is easy and the optimal threshold for parameter pruning is self-determined by the proposed approach employing a validation set.
LOBSTER achieves competitive results from shallow architectures like LeNet-300 and LeNet-5 to deeper topologies like ResNet over ImageNet.
In these scenarios we have observed the boost provided by the proposed regularization approach towards less-unaware approaches like $\ell_2$ regularization, in terms of achieved sparsity.\\
Future research includes the extension of LOBSTER to achieve sparsity with a structure and a thorough evaluation of the savings in terms of memory footprint.

\bibliography{bibliography}

\end{document}